\documentclass[11pt]{article}

\usepackage[preprint]{acl}
\usepackage{microtype}
\usepackage{booktabs}
\usepackage{multirow}
\usepackage{amsmath,amssymb}
\usepackage{graphicx}
\usepackage{enumitem}
\usepackage[ruled]{algorithm2e}
\usepackage{dblfloatfix}
\usepackage{float}
\usepackage{times}
\usepackage{latexsym}
\usepackage[T1]{fontenc}
\usepackage[utf8]{inputenc}
\usepackage{array}
\usepackage[table]{xcolor}
\definecolor{bestgreen}{RGB}{232,245,233}

\title{EVAR: Evidence-Validated Hypothesis Admission for Budget-Aware Narrative Reasoning}

\author{
  Peilin Liu$^{1}$, Zhiquan Ji$^{1}$\thanks{Corresponding author.}, Jinglong Ping$^{2}$ \\
  $^{1}$College of Software, Jilin University \\
  $^{2}$School of Urban Planning and Design, Peking University \\
  \texttt{\{liupl9922,jizq1622\}@mails.jlu.edu.cn} \\
  \texttt{pingjinglong26@stu.pku.edu.cn}
}

\begin{document}
\maketitle

\begin{abstract}
Large language models (LLMs) often produce fluent but weakly grounded conclusions when reasoning over non-interactive, long-form narratives. A central failure mode is that unsupported intermediate hypotheses can enter the reasoning trajectory and contaminate subsequent inference, especially when evidence is scattered across distant parts of the story. To address this problem, we propose EVAR, an evidence-validated hypothesis admission framework for budget-aware narrative reasoning. EVAR first compiles the narrative into an immutable evidence store of source-linked atomic claims and assigns an instance-specific inference budget from unresolved gaps and uncertainty signals. During refinement, EVAR directly proposes candidate hypotheses for unresolved gaps, constructs hypothesis-conditioned validation challenges, and verifies each candidate against the locked store before admission: supported hypotheses enter the answer-supporting state, unverifiable ones are quarantined, and contradictory ones are discarded. A sufficiency-based stopping mechanism further avoids unnecessary refinement. Experiments on NarraCrime and multiple public reasoning benchmarks show that EVAR improves both task performance and evidence faithfulness while maintaining controllable inference cost. Our code is available at
\href{https://github.com/Cosinecos/EVAR}{\textsc{EVAR}}.
\end{abstract}

\section{Introduction}

Large language models excel at localized multi-step reasoning but degrade sharply on non-interactive long-form narratives, where decisive premises are scattered across disjointed events and must be inferred from static text without recourse to clarification or external knowledge~\cite{Liu2023LostMiddle,Kocisky2018NarrativeQA_TACL,Pang2022QuALITY}. This setting exposes two failure modes that conventional benchmarks obscure.

First, premature commitment: models may form an early judgment from incomplete premises and subsequently rationalize it, yielding coherent but unfaithful explanations~\cite{Turpin2023UnfaithfulCoT,Fan2026SABA}.

Second, uniform reasoning depth: indiscriminate self-revision adds
inference cost and may fail to improve---or even degrade---reasoning
accuracy when reliable feedback is unavailable~\cite{Huang2024CannotSelfCorrect,Stechly2023GPT4Wrong}.

We propose EVAR, an evidence-validated hypothesis admission framework for budget-aware narrative reasoning. EVAR first compiles the narrative into an immutable evidence store of source-linked atomic claims with local uncertainty tags, then estimates instance difficulty from unresolved gaps and uncertainty signals to assign an instance-specific inference budget: easy cases are synthesized directly, while hard cases enter verifier-gated refinement.

During refinement, EVAR first proposes candidate hypotheses for
unresolved gaps and then constructs hypothesis-conditioned validation
challenges; each candidate is checked against the evidence store before state update---supported ones admitted, unverifiable ones quarantined, contradictory ones discarded---with early stopping triggered once evidence suffices.

We benchmark EVAR on HotpotQA~\cite{Yang2018HotpotQA}, StrategyQA~\cite{Geva2021StrategyQA}, BBH~\cite{Suzgun2022BBH}, and our new NarraCrime dataset—300 cases across Easy/Medium/Complex splits that stress-test narrative reasoning under increasing story length, evidence density, and suspect complexity. Beyond final-verdict quality, we evaluate both the recovery of required task content and the faithfulness of atomic claims in the final answer.

\noindent\textbf{Our main contributions are as follows:}
\begin{itemize}[leftmargin=*,itemsep=2pt]
\item We formulate evidence-grounded narrative reasoning as a hypothesis admission problem: intermediate hypotheses must be explicitly supported by source evidence before they can affect answer synthesis.
\item We propose EVAR, a budget-aware test-time reasoning framework that builds a provenance-preserving store of source-linked atomic evidence units, assigns instance-specific refinement budgets, and uses hypothesis-conditioned validation challenges to gate state updates—admitting only supported hypotheses while quarantining unverifiable ones and discarding contradictions.
\item We release NarraCrime, a 300-case long-form narrative reasoning benchmark with difficulty-controlled splits and evidence-oriented annotations. Its evaluation combines four task-quality measures—Role-Aware Verdict Score (RVS), Intent Recall (IR), Action Schema Recall (ASR), and Evidence Coverage (EC)—with two final-output, atomic-claim-level faithfulness diagnostics—Unsupported Claim Rate (UCR) and Contradiction Rate (CR).
\end{itemize}

\section{Related Work}
\label{sec:related}

\subsection{Multi-step Prompting and Sampling-based Aggregation}

Chain-of-Thought~\cite{Wei2022CoT}, Self-Consistency~\cite{Wang2023SelfConsistency_ICLR},
and Tree-of-Thoughts~\cite{Yao2023TreeOfThoughts} expand the reasoning
space. However, a larger reasoning space does not guarantee faithful
inference: CoT explanations can rationalize biased answers, while
self-correction without reliable feedback may fail or degrade
performance~\cite{Turpin2023UnfaithfulCoT,Huang2024CannotSelfCorrect,Stechly2023GPT4Wrong}.

\subsection{Iterative Self-feedback and Revision}

Self-Refine and Reflexion improve outputs through iterative
revision cycles~\cite{Madaan2023SelfRefine,Shinn2023Reflexion}.
However, self-revision without reliable external feedback does
not consistently correct factual errors and also requires
additional inference~\cite{Huang2024CannotSelfCorrect,
Stechly2023GPT4Wrong}. Evidence-aware revision methods such as
RARR instead retrieve external sources and post-edit outputs to
improve attribution~\cite{Gao2023RARR}. EVAR targets a different,
closed-world setting in which external retrieval is not used:
candidate hypotheses are validated only against the locked
narrative evidence store before they may update the
answer-supporting state.

\subsection{Structured Reasoning and Trajectory-based Search}

Structured frameworks such as CRITIC, $S^2$R, and
SELF-DISCOVER~\cite{Gou2024CRITIC,Ma2025S2R,Zhou2024SelfDiscover}
move beyond single-pass inference through critique, decomposition,
and strategy discovery, while Graph-of-Thoughts~\cite{Besta2024GoT}
represents reasoning as a graph whose states can be expanded and
aggregated. In long-form narratives, however, unsupported
intermediate conclusions can propagate into later reasoning
steps~\cite{Turpin2023UnfaithfulCoT,Huang2024CannotSelfCorrect},
and exploring or aggregating multiple trajectories increases
inference cost~\cite{Wang2023SelfConsistency_ICLR,Besta2024GoT}.
EVAR therefore validates each candidate against an immutable
narrative evidence store before allowing it to update the
answer-supporting state, reducing the propagation of unsupported
hypotheses while controlling the refinement budget.

SABA~\cite{Fan2026SABA} performs proactive premise-sufficiency assessment before final synthesis by constructing an event--attribute-aligned base state and using queries to generate hypotheses. EVAR centers on evidence-validated admission under instance-specific budgets: it retains source-linked atomic evidence units and applies hypothesis-conditioned validation challenges before each state update, allowing only supported hypotheses to enter the answer-supporting state.

\subsection{Narrative and Reasoning Benchmarks}

Prior work evaluates abductive inference and narrative or social-deduction reasoning~\cite{DelFishel2023TrueDetective,Wu2024DecipheringDetectives,Wang2023aAvalon,Song2025BeyondSurvival,Zhu2024PlayerStar,Zhang2024CollaborationMechanisms}. These works mainly serve as testing grounds, whereas EVAR focuses on reliable, budget-aware test-time reasoning over non-interactive narratives.

\begin{figure*}[t]
  \centering
  \includegraphics[width=1\textwidth,trim=12 10 12 10,clip]{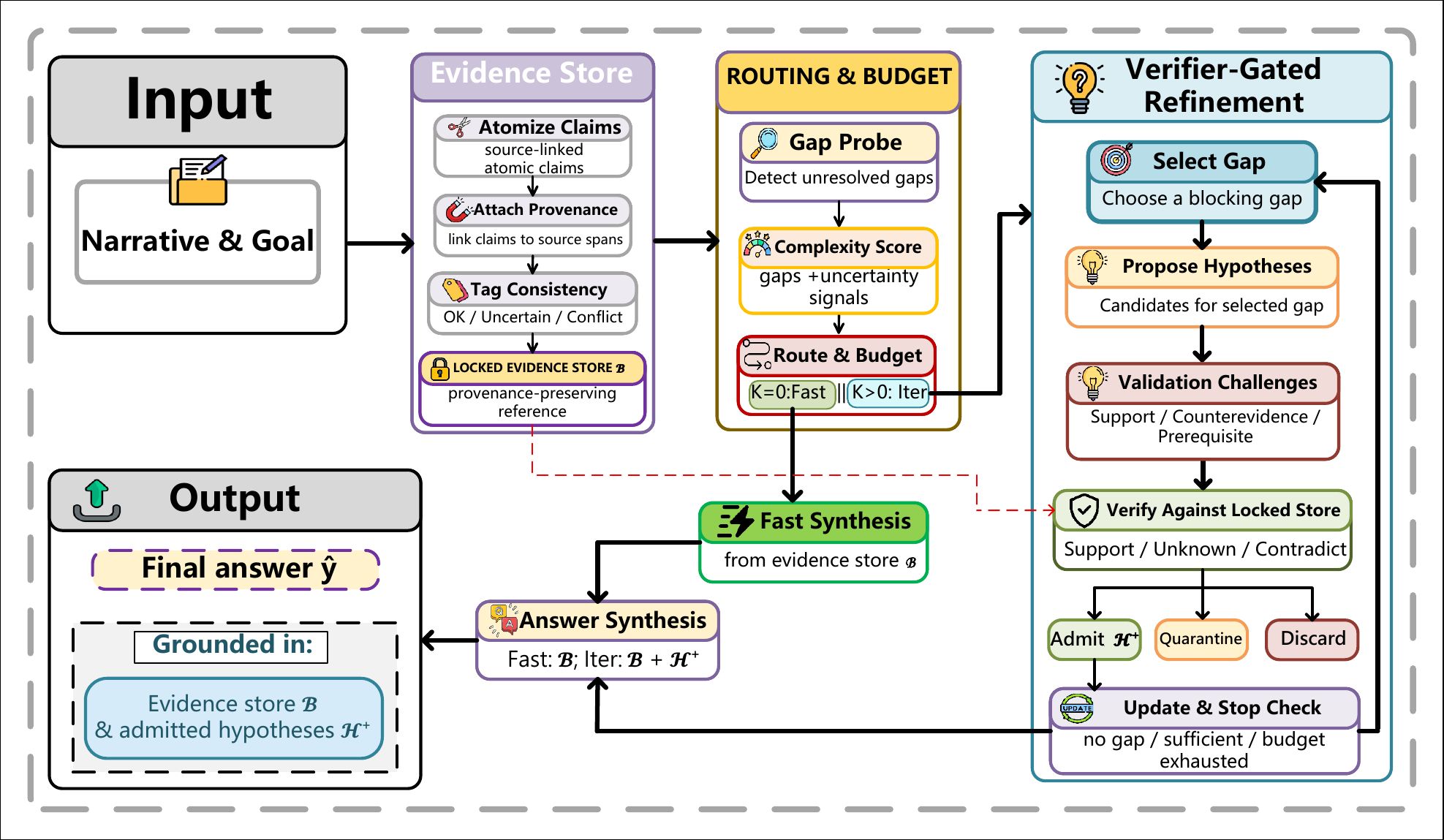}
  \caption{Overview of EVAR. Given the input narrative and goal, EVAR first constructs a locked evidence store and then assigns an instance-specific refinement budget. Easy instances are handled by fast synthesis, while difficult instances enter verifier-gated refinement. Candidate hypotheses are verified against the evidence store: supported hypotheses are admitted into the answer-supporting state, unverifiable hypotheses are quarantined, and contradictory hypotheses are discarded. The final answer is synthesized only from the evidence store and admitted hypotheses.}
  \label{fig:framework}
\end{figure*}

\section{Method}

\subsection{Task Definition and Overview}

We study non-interactive long-form narrative reasoning: given a fixed narrative
\begin{equation}
\mathcal{X}=\{x_i\}_{i=1}^{n},
\end{equation}
where each $x_i$ denotes a sentence or source span. The goal \(\mathcal{G}\) is to infer a conclusion grounded solely in
\(\mathcal{X}\); follow-up questions and external retrieval are outside
our task definition. Such closed-world reasoning still requires models
to locate and integrate dispersed information from long
contexts~\cite{Liu2023LostMiddle}, while unsupported generation and
compositional failures remain risks~\cite{Ji2023HallucinationSurvey,Dziri2023FaithFate}.

We model the LLM as an operator $\mathcal{M}_{\pi}(\cdot)$, where $\pi$ denotes a structured interface enforced through prompting and constrained outputs. EVAR follows an evidence-first strategy: it first compiles $\mathcal{X}$ into a stable evidence store, then applies budgeted refinement only to hard instances, admitting new hypotheses only after evidence validation. The framework consists of evidence store construction, budget-aware routing, and verifier-gated hypothesis admission. Thus, refinement never inserts an unverified hypothesis into the answer-supporting state; unsupported, unverifiable, or contradictory hypotheses are blocked from answer synthesis.

\subsection{Evidence Store Construction}

EVAR first atomizes the narrative into source-grounded evidence claims:
\begin{equation}
\{(c_j,\rho_j)\}_{j=1}^{m}
\leftarrow
\mathcal{M}_{\pi_{\textsc{atom}}}(\mathcal{X}),
\end{equation}
where \(c_j\) is an atomic proposition explicitly expressed in the narrative and \(\rho_j\subseteq\{1,\ldots,n\}\) records the sentence or span identifiers from which it is extracted. Each claim must retain at least one valid source identifier; inferred motives or causal bridges are excluded at this stage.

We then attach only source-observable metadata without merging claims into reconstructed events:
\begin{equation}
\begin{aligned}
u_j &= (c_j,\rho_j,\eta_j),\\
\eta_j &= (\textsf{entity}_j,\textsf{time}_j,\textsf{polarity}_j),\\
\mathcal{U} &= \{u_j\}_{j=1}^{m}.
\end{aligned}
\end{equation}
Here, \(u_j\) is a provenance-preserving evidence unit and \(\eta_j\) contains lightweight metadata explicitly recoverable from its cited spans. This construction keeps non-contiguous claims separate and preserves their original textual provenance.

Each evidence unit then receives a localized consistency tag:
\begin{equation}
\begin{aligned}
\kappa_j &\leftarrow \mathcal{M}_{\pi_{\textsc{tag}}}(u_j,\mathcal{U}\setminus\{u_j\}),\\
\kappa_j &= (\textsf{status}_j,\textsf{sev}_j,\textsf{note}_j),
\end{aligned}
\end{equation}
where
\begin{equation}
\begin{aligned}
\textsf{status}_j &\in \{\textsf{OK},\textsf{Uncertain},\textsf{Conflict}\},\\
\textsf{sev}_j &\in \{0,1,2,3\}.
\end{aligned}
\end{equation}
The status indicates whether the unit is normal, uncertain, or conflicting, while the severity score records how serious the issue is. The resulting immutable base evidence store is
\begin{equation}
\mathcal{B}=\{(u_j,\kappa_j)\}_{j=1}^{m}.
\end{equation}
This store serves as the stable reference for all downstream reasoning.

\subsection{Budget-aware Routing and Hypothesis Admission}

Given \(\mathcal{B}\), EVAR first performs a lightweight gap probe:
\begin{equation}
\mathcal{Z}_0 \leftarrow \mathcal{M}_{\pi_{\textsc{gap}}}(\mathcal{B},\mathcal{G}),
\end{equation}
where each element in \(\mathcal{Z}_0\) represents an underspecified premise that currently blocks a reliable conclusion. Based on these gaps and the tagged evidence units, EVAR computes a simple complexity score:
\begin{equation}
\Gamma = \alpha_1|\mathcal{Z}_0|
+ \alpha_2 \sum_{j=1}^{m}\mathbb{I}\{\textsf{status}_j \neq \textsf{OK}\}
+ \alpha_3 \sum_{j=1}^{m} \textsf{sev}_j.
\label{eq:complexity}
\end{equation}
This score reflects three sources of difficulty: how many unresolved evidence gaps remain, how many evidence units are uncertain or conflicting, and how severe these local evidence issues are.

EVAR converts the complexity score into an explicit instance-specific refinement budget. 
Instead of asking the model to freely decide how much to reason, we compute the budget under a global cap \(B_{\max}\):
\begin{equation}
\begin{aligned}
K
=
\min\Bigg(
B_{\max},
\max\Big(
0,
\Big\lceil
\frac{\Gamma-\tau_{\mathrm{fast}}}
{\tau_{\mathrm{step}}}
\Big\rceil
\Big)
\Bigg).
\end{aligned}
\label{eq:budget}
\end{equation}
The route is then determined by the assigned budget:
\begin{equation}
r=
\begin{cases}
\textsc{Fast}, & K=0,\\
\textsc{Iter}, & K>0.
\end{cases}
\label{eq:route}
\end{equation}
Here, \(\tau_{\mathrm{fast}}\) controls when an instance can bypass refinement, and \(\tau_{\mathrm{step}}\) controls how quickly the refinement budget increases with estimated complexity.

During refinement, EVAR maintains a working refinement state
\begin{equation}
\mathcal{S}_t =
(\mathcal{B},\mathcal{V}_{\textsc{hist},t},\mathcal{H}^{+}_{\textsc{hist},t}),
\end{equation}
where \(\mathcal{V}_{\textsc{hist},t}\) denotes the accumulated hypothesis-conditioned validation challenges and \(\mathcal{H}^{+}_{\textsc{hist},t}\) denotes the evidence-supported hypotheses admitted before iteration \(t\). The challenge history is retained for audit and duplicate avoidance, but final answer synthesis is restricted to \(\mathcal{B}\) and \(\mathcal{H}^{+}_{\textsc{hist},t}\). Unverifiable and contradictory hypotheses are logged separately and are not included in the answer-supporting state.

At iteration \(t\), EVAR identifies the remaining gaps
\begin{equation}
\mathcal{Z}_t \leftarrow \mathcal{M}_{\pi_{\textsc{gap}}}(\mathcal{S}_t,\mathcal{G}),
\end{equation}
and directly proposes candidate hypotheses for each unresolved gap:
\begin{equation}
\begin{aligned}
\mathcal{H}_{z,t} &\leftarrow \mathcal{M}_{\pi_{\textsc{hyp}}}(z,\mathcal{S}_t),\\
\mathcal{H}_t &= \bigcup_{z\in\mathcal{Z}_t}\mathcal{H}_{z,t}.
\end{aligned}
\end{equation}
For each candidate, EVAR constructs hypothesis-conditioned validation challenges:
\begin{equation}
\begin{aligned}
\mathcal{V}_{h,t}
&\leftarrow \mathcal{M}_{\pi_{\textsc{chal}}}(h,\mathcal{B}),\\
\mathcal{V}_{h,t}
&=
\{v^{\mathrm{sup}}_{h,t},v^{\mathrm{ctr}}_{h,t},v^{\mathrm{req}}_{h,t}\},\\
\mathcal{V}_t &= \bigcup_{h\in\mathcal{H}_t}\mathcal{V}_{h,t}.
\end{aligned}
\end{equation}
Here, $v^{\mathrm{sup}}_{h,t}$, $v^{\mathrm{ctr}}_{h,t}$, and $v^{\mathrm{req}}_{h,t}$ respectively request direct supporting evidence, counterevidence, and indispensable premises that remain unsupported. These challenges are not treated as evidence or hypotheses and are never added to the answer-supporting component; they are retained only for audit and duplicate avoidance.

To prevent unsupported content from entering the reasoning state, each candidate hypothesis is verified against the locked evidence store:
\begin{equation}
\begin{aligned}
\bigl(\ell(h),\mathcal{E}_h\bigr)
&\leftarrow \mathcal{M}_{\pi_{\textsc{ver}}}(h,\mathcal{V}_{h,t},\mathcal{B}),\\
\ell(h) &\in \{\textsf{Support},\textsf{Unknown},\textsf{Contradict}\}.
\end{aligned}
\label{eq:ver_label}
\end{equation}
EVAR then applies a strict hypothesis admission rule. Only hypotheses labeled as \textsf{Support} are admitted into the answer-supporting component:
\begin{equation}
\mathcal{H}^{+}_t
=
\{(h,\mathcal{E}_h)
\mid
h\in\mathcal{H}_t,\;
\ell(h)=\textsf{Support}
\},
\label{eq:supported_hyp}
\end{equation}
where \(\mathcal{E}_h\subseteq\mathcal{B}\) denotes the supporting evidence units for \(h\). Hypotheses labeled as \textsf{Unknown} are placed into a quarantine buffer and are not allowed to affect final answer synthesis:
\begin{equation}
\mathcal{H}^{?}_t
=
\{h
\mid
h\in\mathcal{H}_t,\;
\ell(h)=\textsf{Unknown}
\}.
\label{eq:unknown_hyp}
\end{equation}
Hypotheses labeled as \textsf{Contradict} are discarded:
\begin{equation}
\mathcal{H}^{-}_t
=
\{h
\mid
h\in\mathcal{H}_t,\;
\ell(h)=\textsf{Contradict}
\}.
\label{eq:contra_hyp}
\end{equation}
The working state records validation challenges for audit and duplicate avoidance, while its answer-supporting component is updated only with evidence-supported hypotheses:
\begin{equation}
\begin{aligned}
\mathcal{S}_{t+1}
\leftarrow
\bigl(
&\mathcal{B},
\mathcal{V}_{\textsc{hist},t}\cup \mathcal{V}_t,\\
&\mathcal{H}^{+}_{\textsc{hist},t}\cup \mathcal{H}^{+}_t
\bigr).
\end{aligned}
\label{eq:admission_update}
\end{equation}

Refinement stops when no blocking gap remains, i.e., \(\mathcal{Z}_t=\emptyset\), or when the current state is already sufficient:
\begin{equation}
\begin{aligned}
\sigma_{t+1} &\leftarrow \mathcal{M}_{\pi_{\textsc{suf}}}(\mathcal{S}_{t+1},\mathcal{G}),\\
\sigma_{t+1} &\ge \tau_{\textsc{suf}}.
\end{aligned}
\end{equation}
Let \(t^\star\) denote the index of the terminal refinement state reached when the refinement loop stops, and let \(\mathcal{S}_{t^\star}\) denote that state. To prevent validation challenges from leaking into answer generation, EVAR forms an answer-supporting terminal state that contains only the locked evidence store and admitted hypotheses:
\begin{equation}
\begin{aligned}
\mathcal{S}^{\mathrm{ans}}_{t^\star}
&=
(\mathcal{B},\mathcal{H}^{+}_{\textsc{hist},t^\star}),\\
\bigl(\hat{y},\mathbf{p}\bigr)
&\leftarrow
\mathcal{M}_{\pi_{\textsc{ans}}}(\mathcal{S}^{\mathrm{ans}}_{t^\star},\mathcal{G})
\qquad \text{(NarraCrime)}.
\end{aligned}
\end{equation}
For the public benchmarks, only the textual answer \(\hat{y}\) is requested and evaluated.

\section{Experiments}
\label{sec:exp}

\subsection{Experimental Setup}
\label{sec:exp_setup}

\paragraph{Datasets.}
We evaluate EVAR on one non-interactive long-form narrative benchmark, NarraCrime (statistics in Table~\ref{tab:dp_stats}), and three public reasoning benchmarks: HotpotQA (HQA), StrategyQA (SQA), and BBH~\cite{Yang2018HotpotQA,Geva2021StrategyQA,Suzgun2022BBH}. For StrategyQA, all methods are evaluated on all 2,780 examples in the official dataset rather than only on the commonly used development split. For NarraCrime, we report results on the \textit{Easy/Medium/Complex} splits to reflect increasing levels of story length, evidence density, and cross-event reasoning complexity. Each case contains exactly one principal culprit and zero or more confirmed accomplices. For notational uniformity, the principal culprit is represented as a singleton set, and the culprit and accomplice sets are disjoint.

\paragraph{Baselines.}
We compare EVAR with representative test-time reasoning baselines, including Direct, CoT, Self-Refine, SC, CRITIC, $S^2$R-style, SELF-DISC., and GoT. Here, $S^2$R-style denotes a prompt-only implementation of the self-verification and self-correction procedure in Ma et al.~(2025) under the shared DeepSeek-V3.2 backbone, without reproducing its original SFT/RL training pipeline. These baselines correspond to different reasoning paradigms, including direct answering, chain-of-thought reasoning, self-feedback refinement, sampling-based aggregation, structured reflection, and trajectory-based search.

\paragraph{Implementation Details.}
Unless otherwise specified, all main experiments use DeepSeek-V3.2 as the backbone model. All compared methods receive the same input and follow a unified experimental protocol. For NarraCrime, every method's final call follows the same structured output schema, containing a textual answer and a candidate-ID--probability map over the fixed candidate set; the evaluator requires complete candidate coverage, non-negative probabilities, and unit total mass within a fixed numerical tolerance. The exact schema and malformed-output handling are provided in the released evaluator, and no provider-specific token log-probabilities are used. We use greedy decoding with temperature=0, top-$p$=1.0, and a maximum output length of 512 tokens. For stochastic baselines, we enable sampling and use a non-zero temperature while keeping the other settings unchanged. All reported results are averaged over three independent runs. For stochastic decoding, we use random seeds 42, 44, and 46. Table~\ref{tab:main_results} reports task metrics as mean$\pm$std. For compactness, $T$ and the scalar results in Tables~\ref{tab:ablation_evar_sorted}--\ref{tab:routing_public} and Figs.~\ref{fig:cost_utility}--\ref{fig:earlystop_real} are reported as means over the same three runs, with standard deviations omitted. To examine whether the gains of EVAR generalize beyond the main backbone, we additionally report backbone robustness results with several representative additional backbone models in Appendix~\ref{app:backbone_robustness}.
\begin{table}[t]
\centering
\small
\setlength{\tabcolsep}{5pt}
\renewcommand{\arraystretch}{1.08}
\resizebox{\columnwidth}{!}{
\begin{tabular}{l c c c c}
\toprule
Split & \#Cases & Words & Cues & Suspects \\
\midrule
Easy    & 100 & 863.5  & 8.7  & 3.6 \\
Medium  & 100 & 1065.2 & 10.9 & 4.0 \\
Complex & 100 & 1413.4 & 13.4 & 4.5 \\
\midrule
Total   & 300 & 1114.1 & 11.0 & 4.0 \\
\bottomrule
\end{tabular}
}
\caption{Dataset statistics of NarraCrime. Words, Cues, and Suspects denote average story length, average number of annotated evidence cues, and average number of candidate suspects, respectively.}
\label{tab:dp_stats}
\end{table}

\subsection{Metrics}
\label{sec:exp_baselines_metrics}

\paragraph{Metric Definition.}
For the public benchmarks, we report Answer exact match (Ans) and Supporting-Fact F1 (SF) on HotpotQA, accuracy on StrategyQA, and macro-averaged accuracy across all 23 BBH tasks. We report four task-quality metrics on NarraCrime: Role-Aware Verdict Score (RVS), Intent Recall (IR), Action Schema Recall (ASR), and Evidence Coverage (EC). RVS evaluates final-verdict
quality while distinguishing principal culprits from confirmed
accomplices. For instance \(i\), let \(\Omega_i\) denote the fixed
set of candidate suspects, and let
\(\mathcal{R}^{\mathrm{cul}}_i\) and
\(\mathcal{R}^{\mathrm{acc}}_i\) denote the disjoint gold sets of
principal culprits and confirmed accomplices, respectively. Each
method returns a normalized probability distribution
\(\mathbf{p}_i=\{p_i(e):e\in\Omega_i\}\), where
\(p_i(e)\geq 0\) and \(\sum_{e\in\Omega_i}p_i(e)=1\). We define
\begin{equation}
\begin{aligned}
s_i(\mathbf{p}_i)
&=
\sum_{e\in\mathcal{R}^{\mathrm{cul}}_i}p_i(e)
+
\lambda\sum_{e\in\mathcal{R}^{\mathrm{acc}}_i}p_i(e),\\
\mathrm{RVS}
&=
\frac{1}{N}\sum_{i=1}^{N}s_i(\mathbf{p}_i),
\qquad \lambda=0.5.
\end{aligned}
\label{eq:rvs}
\end{equation}

Probability mass assigned to a principal culprit receives full credit, mass assigned to a confirmed accomplice receives partial credit because it captures part of the responsible group, and mass assigned to other candidates receives no credit. Because the distribution has unit total mass, RVS rewards probability mass assigned to the gold roles rather than the number of candidates assigned nonzero probability. We fix \(\lambda=0.5\) for all reported methods, splits, and backbones; it is not tuned separately for any method or dataset split. RVS is computed directly from the reported probability distribution and the gold role annotations without LLM-based adjudication. IR and ASR measure the recovery of the annotated intent and action schema, respectively, while EC measures the coverage of annotated supporting evidence. We instantiate these three content-recovery dimensions using the unified proposition-level semantic-matching protocol described below.

For $X\in\{I,A,E\}$, we compute recall based on semantic matching between the predicted proposition set $\hat{\mathcal{P}}^{X}$ and the reference proposition set $\mathcal{P}^{X}$; additional implementation details are provided in Appendix~\ref{app:metric_details}. Specifically, we first compute cosine similarity between propositions using the encoder $\phi(\cdot)$:
\begin{equation}
\mathrm{sim}(\hat{p},p)=\frac{\phi(\hat{p})^\top \phi(p)}{\|\phi(\hat{p})\|_2\,\| \phi(p)\|_2},
\label{eq:sim}
\end{equation}
We then perform greedy one-to-one matching under a threshold and normalize the number of matched reference propositions into recall:
\begin{equation}
\mathrm{Recall}(\hat{\mathcal{P}}^{X},\mathcal{P}^{X})
=
\frac{\mathrm{match}(\hat{\mathcal{P}}^{X},\mathcal{P}^{X})}{|\mathcal{P}^{X}|}.
\label{eq:recall}
\end{equation}

Using this common recall formulation, we report the three content-recovery metrics as follows:
\begin{equation}
\mathrm{IR}=\mathrm{Recall}(\hat{\mathcal{P}}^{I},\mathcal{P}^{I}).
\label{eq:ir}
\end{equation}
\begin{equation}
\mathrm{ASR}=\mathrm{Recall}(\hat{\mathcal{P}}^{A},\mathcal{P}^{A}).
\label{eq:asr}
\end{equation}
\begin{equation}
\mathrm{EC}=\mathrm{Recall}(\hat{\mathcal{P}}^{E},\mathcal{P}^{E}).
\label{eq:ec}
\end{equation}

\paragraph{LLM-based Evaluation.}
All LLM-based evaluation operations---predicted-proposition
extraction, atomic-claim decomposition, and evidence-status
labeling---are performed automatically by GPT-5.5
(\texttt{gpt-5.5}) using fixed task-specific prompts and a common
decoding configuration across all evaluated systems. For
proposition extraction and atomic decomposition, the judge
receives the generated prediction but not the identity of the
evaluated method; for evidence-status labeling, it additionally
receives the gold evidence annotations. No human adjudication is
used. For NarraCrime, the judge uses only the textual answer; the
probability field is excluded from IR, ASR, EC, UCR, and CR.

Following prior claim-level factuality evaluation~\cite{Akbar2024HalluMeasure}, we report two final-output faithfulness diagnostics for NarraCrime at the atomic-claim level: Unsupported Claim Rate (UCR) and Contradiction Rate (CR). Unlike the preceding content-recovery metrics, these diagnostics evaluate factual additions in the model's final prediction rather than the recovery of reference propositions. We first decompose each final prediction into a set of atomic claims \(\hat{\mathcal{C}}_i\). Each claim is independently judged against the gold evidence annotations and assigned one of three labels: \textsf{Support}, \textsf{Unknown}, or \textsf{Contradict}. UCR measures the fraction of final-output claims that are not supported by the annotated evidence, including both unverifiable and contradictory claims:
\begin{equation}
\mathrm{UCR}
=
\frac{
\sum_{i=1}^{N}
\left|
\left\{
c\in\hat{\mathcal{C}}_i
\;\middle|\;
\substack{
\ell(c)\in
\{\textsf{Unknown},\\[-1pt]
\textsf{Contradict}\}
}
\right\}
\right|
}{
\sum_{i=1}^{N}|\hat{\mathcal{C}}_i|
}.
\label{eq:ucr}
\end{equation}
CR measures the fraction of claims that contradict the annotated evidence:
\begin{equation}
\mathrm{CR}
=
\frac{
\sum_{i=1}^{N}
\left|
\{c\in\hat{\mathcal{C}}_i \mid \ell(c)=\textsf{Contradict}\}
\right|
}{
\sum_{i=1}^{N}|\hat{\mathcal{C}}_i|
}.
\label{eq:cr}
\end{equation}
Because every contradictory claim is also unsupported, claims labeled as \textsf{Contradict} contribute to both UCR and CR, whereas claims labeled as \textsf{Unknown} contribute only to UCR. Thus, the claims counted by CR form a subset of those counted by
UCR, so \(0\leq\mathrm{CR}\leq\mathrm{UCR}\); the two reported
rates are intentionally not mutually exclusive.

Lower UCR and CR indicate stronger evidence faithfulness. All six metrics are defined on the interval $[0,1]$. For readability, all tables report $100\times$ these values on a $0$--$100$ scale.

\subsection{EVAR Inference Configuration}
\label{sec:exp_impl}

We measure inference cost $T$ as the average number of LLM calls per instance. Each invocation of an EVAR operator $\mathcal{M}_{\pi}(\cdot)$ is counted as one call. The global budget cap $B_{\max}$ bounds the instance-specific refinement budget $K$, but $T$ is not strictly linear in $K$ because the numbers of gaps, candidate hypotheses, and validation challenges vary across instances, and early stopping may terminate refinement before the budget is exhausted.

\subsection{Main Results}
\label{sec:exp_main}

\begin{table*}[!t]
\centering
\scriptsize
\setlength{\tabcolsep}{4.0pt}
\renewcommand{\arraystretch}{1.10}

\resizebox{0.98\textwidth}{!}{%
\begin{tabular}{l l c c c c c c c c c}
\toprule
\multirow{2}{*}{Method} & \multirow{2}{*}{Split} & \multicolumn{4}{c}{NarraCrime} & \multicolumn{2}{c}{HQA} & SQA & BBH & \multirow{2}{*}{$T$} \\
\cmidrule(lr){3-6} \cmidrule(lr){7-8} \cmidrule(lr){9-9} \cmidrule(lr){10-10}
& & RVS & IR & ASR & EC & Ans & SF & Acc & Acc & \\
\midrule

\multirow{3}{*}{Direct}
& Easy     & $65.6\pm0.54$ & $64.8\pm0.38$ & $60.7\pm0.61$ & $70.8\pm0.47$ &
\multirow{3}{*}{$62.1\pm0.63$} & \multirow{3}{*}{$52.6\pm0.58$} &
\multirow{3}{*}{$82.7\pm0.41$} & \multirow{3}{*}{$79.0\pm0.46$} &
\multirow{3}{*}{$\mathbf{1.0}$} \\
& Medium & $48.7\pm0.66$ & $63.6\pm0.52$ & $58.6\pm0.57$ & $61.9\pm0.60$ \\
& Complex   & $40.1\pm0.69$ & $60.0\pm0.63$ & $58.2\pm0.65$ & $59.4\pm0.67$ \\
\midrule

\multirow{3}{*}{CoT}
& Easy     & $67.7\pm0.43$ & $77.9\pm0.31$ & $64.6\pm0.49$ & $76.2\pm0.36$ &
\multirow{3}{*}{$68.9\pm0.54$} & \multirow{3}{*}{$60.0\pm0.50$} &
\multirow{3}{*}{$86.9\pm0.33$} & \multirow{3}{*}{$86.6\pm0.39$} &
\multirow{3}{*}{$\underline{2.4}$} \\
& Medium & $58.0\pm0.58$ & $65.7\pm0.44$ & $62.3\pm0.55$ & $71.6\pm0.60$ \\
& Complex   & $46.0\pm0.62$ & $60.3\pm0.51$ & $60.4\pm0.63$ & $61.6\pm0.56$ \\
\midrule

\multirow{3}{*}{Self-Refine}
& Easy     & $71.1\pm0.39$ & $66.5\pm0.46$ & $68.8\pm0.41$ & $79.7\pm0.44$ &
\multirow{3}{*}{$67.9\pm0.51$} & \multirow{3}{*}{$59.7\pm0.55$} &
\multirow{3}{*}{$87.6\pm0.35$} & \multirow{3}{*}{$88.0\pm0.37$} &
\multirow{3}{*}{$6.2$} \\
& Medium & $65.2\pm0.47$ & $66.9\pm0.42$ & $64.8\pm0.59$ & $76.5\pm0.50$ \\
& Complex   & $54.8\pm0.63$ & $62.7\pm0.57$ & $60.6\pm0.64$ & $66.1\pm0.61$ \\
\midrule

\multirow{3}{*}{SC ($k=5$)}
& Easy     & $71.5\pm0.28$ & $69.1\pm0.37$ & $68.9\pm0.33$ & $81.9\pm0.35$ &
\multirow{3}{*}{$72.7\pm0.44$} & \multirow{3}{*}{$61.6\pm0.48$} &
\multirow{3}{*}{$90.8\pm0.29$} & \multirow{3}{*}{$88.9\pm0.32$} &
\multirow{3}{*}{$12.1$} \\
& Medium & $70.7\pm0.41$ & $64.2\pm0.49$ & $66.8\pm0.52$ & $79.8\pm0.46$ \\
& Complex   & $62.4\pm0.55$ & $62.3\pm0.60$ & $61.9\pm0.58$ & $71.4\pm0.57$ \\
\midrule

\multirow{3}{*}{CRITIC}
& Easy     & $77.9\pm0.26$ & $67.6\pm0.35$ & $71.2\pm0.38$ & $87.9\pm0.27$ &
\multirow{3}{*}{$74.6\pm0.39$} & \multirow{3}{*}{$68.0\pm0.42$} &
\multirow{3}{*}{$91.1\pm0.25$} & \multirow{3}{*}{$87.3\pm0.33$} &
\multirow{3}{*}{$29.2$} \\
& Medium & $73.2\pm0.36$ & $65.9\pm0.41$ & $68.1\pm0.49$ & $\underline{88.3}\pm0.34$ \\
& Complex   & $65.6\pm0.47$ & $65.2\pm0.52$ & $63.8\pm0.57$ & $72.7\pm0.50$ \\
\midrule

\multirow{3}{*}{$S^2R$-style}
& Easy     & $80.1\pm0.22$ & $69.9\pm0.29$ & $77.1\pm0.35$ & $90.0\pm0.24$ &
\multirow{3}{*}{$72.9\pm0.33$} & \multirow{3}{*}{$68.9\pm0.38$} &
\multirow{3}{*}{$92.4\pm0.21$} & \multirow{3}{*}{$89.9\pm0.27$} &
\multirow{3}{*}{$18.5$} \\
& Medium & $75.6\pm0.31$ & $67.4\pm0.34$ & $71.7\pm0.46$ & $87.7\pm0.32$ \\
& Complex   & $69.0\pm0.44$ & $62.9\pm0.59$ & $\underline{68.0}\pm0.47$ & $77.2\pm0.51$ \\
\midrule

\multirow{3}{*}{SELF-DISC.}
& Easy     & $80.3\pm0.20$ & $72.9\pm0.26$ & $77.8\pm0.33$ & $86.7\pm0.30$ &
\multirow{3}{*}{$76.0\pm0.31$} & \multirow{3}{*}{$67.9\pm0.35$} &
\multirow{3}{*}{$\underline{92.9}\pm0.19$} & \multirow{3}{*}{$\underline{90.7}\pm0.23$} &
\multirow{3}{*}{$4.8$} \\
& Medium & $75.9\pm0.29$ & $67.6\pm0.32$ & $70.6\pm0.45$ & $87.3\pm0.34$ \\
& Complex   & $68.0\pm0.40$ & $65.1\pm0.48$ & $64.6\pm0.53$ & $76.3\pm0.44$ \\
\midrule

\multirow{3}{*}{GoT}
& Easy     & $\underline{83.7}\pm0.17$ & $\underline{78.9}\pm0.21$ & $\underline{81.8}\pm0.24$ & $\underline{90.8}\pm0.20$ &
\multirow{3}{*}{$\underline{76.6}\pm0.25$} & \multirow{3}{*}{$\underline{72.7}\pm0.28$} &
\multirow{3}{*}{$92.2\pm0.18$} & \multirow{3}{*}{$90.0\pm0.22$} &
\multirow{3}{*}{$35.5$} \\
& Medium & $\underline{77.3}\pm0.24$ & $\underline{79.6}\pm0.19$ & $\underline{72.9}\pm0.27$ & $87.6\pm0.23$ \\
& Complex   & $\underline{69.6}\pm0.30$ & $\underline{70.1}\pm0.33$ & $66.7\pm0.39$ & $\underline{77.6}\pm0.31$ \\
\midrule

\rowcolor{bestgreen}
& Easy
& $\mathbf{85.9}\pm0.15$
& $\mathbf{82.7}\pm0.18$
& $\mathbf{85.6}\pm0.16$
& $\mathbf{93.5}\pm0.14$
& & & & & \\

\rowcolor{bestgreen}
& Medium
& $\mathbf{81.6}\pm0.20$
& $\mathbf{80.8}\pm0.19$
& $\mathbf{77.6}\pm0.21$
& $\mathbf{90.6}\pm0.18$
& & & & & \\

\rowcolor{bestgreen}
\multirow{-3}{*}{\textbf{EVAR}}
& Complex
& $\mathbf{78.6}\pm0.24$
& $\mathbf{72.6}\pm0.25$
& $\mathbf{72.2}\pm0.29$
& $\mathbf{83.3}\pm0.23$
& \multirow{-3}{*}{$\mathbf{78.2}\pm0.22$}
& \multirow{-3}{*}{$\mathbf{74.1}\pm0.20$}
& \multirow{-3}{*}{$\mathbf{94.1}\pm0.13$}
& \multirow{-3}{*}{$\mathbf{93.6}\pm0.17$}
& \multirow{-3}{*}{$15.8$} \\

\bottomrule
\end{tabular}%
}

\caption{Main results with DeepSeek-V3.2 as the backbone on NarraCrime (reported by split), HotpotQA (HQA), StrategyQA (SQA), and BBH. Task metrics are reported as mean$\pm$std over three independent runs. \(T\) is reported as the three-run mean number of LLM calls per NarraCrime-Complex instance; lower is better.}
\label{tab:main_results}
\end{table*}
Table~\ref{tab:main_results} summarizes the main results. Overall, EVAR achieves the best performance across all benchmarks while keeping inference cost controllable through routing and budgeted refinement. On NarraCrime, EVAR improves both role-aware verdict quality and task-content recovery metrics across all splits, with the largest gains on NarraCrime-Complex: compared with GoT and SELF-DISC., RVS increases to $78.6$ from $69.6$ and $68.0$, respectively, while EC increases to $83.3$ from $77.6$ and $76.3$. Similar trends also hold on the Easy split, where EVAR reaches $85.9$ RVS and $93.5$ EC.

On public reasoning benchmarks, EVAR also generalizes well. On HQA, it achieves $78.2$ Ans and $74.1$ SF, compared with $76.6$ and $72.7$ from the strongest baseline (GoT), corresponding to relative gains of $2.1\%$ and $1.9\%$, respectively. On SQA and BBH, EVAR reaches $94.1$ and $93.6$, corresponding to relative gains of $1.3\%$ and $3.2\%$. Moreover, on NarraCrime-Complex, compared with GoT ($T=35.5$), EVAR reduces inference cost to $T=15.8$, a $55.5\%$ reduction.

\section{Further Analysis}
\label{sec:analysis}

\subsection{Ablation Study}
\label{sec:exp_ablation}

We conduct controlled ablations on NarraCrime-Complex and StrategyQA to isolate the contribution of each core component in EVAR. All variants use the same backbone model, inputs, prompt templates, decoding settings, and maximum refinement budget as full EVAR; unless otherwise stated, all other settings remain unchanged. Starting from the full pipeline, we remove one component at a time: 
(1) w/o Admission, which disables verifier-gated hypothesis admission and allows generated hypotheses to enter the state without strict support filtering; 
(2) w/o Evidence Store, which removes the locked structured evidence basis; and 
(3) w/o Budget Routing, which removes instance-specific budget assignment.
We also include CoT as a prompting baseline.

As shown in Table~\ref{tab:ablation_evar_sorted}, full EVAR achieves the best overall performance. Removing budget routing causes the largest performance drop, suggesting that instance-specific computation is important for difficult examples. Removing hypothesis admission also leads to a clear increase in unsupported claims, showing that EVAR's gain does not simply come from using more refinement steps. Removing the evidence store further weakens evidence grounding. Overall, the gains of EVAR come from the combination of locked evidence, verifier-gated admission, and budget-aware control.

\begin{table}[t]
\centering
\scriptsize
\setlength{\tabcolsep}{3pt}
\renewcommand{\arraystretch}{1.08}
\begin{tabular*}{\columnwidth}{@{\extracolsep{\fill}}l c c c c@{}}
\toprule
Variant & RVS & EC & UCR$\downarrow$ & SQA \\
\midrule
EVAR              & 78.6 & 83.3 & 8.6  & 94.1 \\
w/o Admission     & 71.5 & 74.4 & 17.8 & 90.5 \\
w/o Evidence Store & 70.3 & 72.5 & 19.4 & 83.2 \\
w/o Budget Routing & 62.9 & 64.7 & 21.7 & 77.5 \\
CoT               & 46.0 & 61.6 & 28.4 & 86.9 \\
\bottomrule
\end{tabular*}
\caption{Ablation study on NarraCrime-Complex and StrategyQA. UCR denotes unsupported claim rate; lower is better.}
\label{tab:ablation_evar_sorted}
\end{table}

\subsection{Hypothesis Admission and Evidence Faithfulness}
\label{sec:faithfulness_analysis}

To examine whether EVAR improves reasoning by controlling which intermediate claims are allowed to affect the final answer, we analyze evidence faithfulness on NarraCrime-Complex. As shown in Table~\ref{tab:faithfulness}, EVAR substantially reduces both unsupported claim rate and contradiction rate compared with strong reasoning baselines. Together with the admission ablation in
Table~\ref{tab:ablation_evar_sorted}, this pattern is consistent
with the intended effect of filtering unverifiable and contradictory
hypotheses before they enter the active reasoning state.

\begin{table}[t]
\centering
\scriptsize
\setlength{\tabcolsep}{3pt}
\renewcommand{\arraystretch}{1.08}
\begin{tabular*}{\columnwidth}{@{\extracolsep{\fill}}l c c c c@{}}
\toprule
Method & RVS & EC & UCR$\downarrow$ & CR$\downarrow$ \\
\midrule
Direct      & 40.1 & 59.4 & 31.6 & 14.2 \\
CoT         & 46.0 & 61.6 & 28.4 & 12.7 \\
Self-Refine & 54.8 & 66.1 & 25.1 & 10.5 \\
GoT         & 69.6 & 77.6 & 15.9 & 7.4 \\
EVAR        & 78.6 & 83.3 & 8.6  & 3.8 \\
\bottomrule
\end{tabular*}
\caption{Evidence-faithfulness analysis on NarraCrime-Complex. UCR and CR denote unsupported claim rate and contradiction rate; lower is better.}
\label{tab:faithfulness}
\end{table}

\subsection{Cost--Utility Trade-off}
\label{sec:exp_cost}

To evaluate controllability, we sweep the global maximum iteration cap $B_{\max}\in\{0,1,2,3,4\}$ while keeping all other settings fixed. We evaluate different budget settings on NarraCrime-Complex and StrategyQA, and measure computation cost by the average number of LLM calls per instance, denoted by $T$.

\begin{figure}[t]
    \centering
    \includegraphics[width=\linewidth]{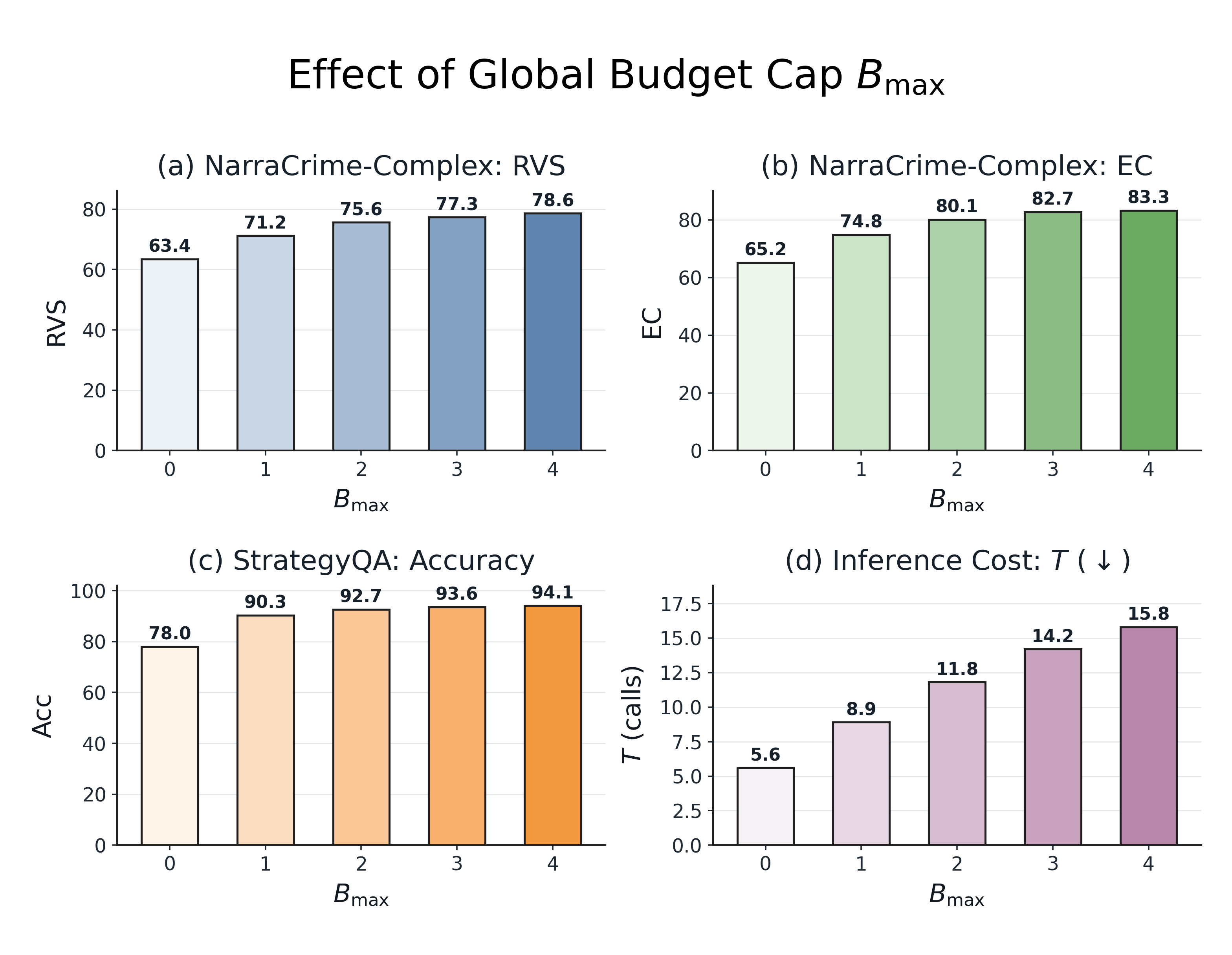}
    \caption{Cost--utility trade-off of EVAR under different global budget caps $B_{\max}$. Panels (a)--(d) report NC-Complex RVS, NC-Complex EC, SQA Acc, and cost $T$, respectively.}
    \label{fig:cost_utility}
\end{figure}

As shown in Fig.~\ref{fig:cost_utility}, EVAR achieves clear performance gains under small budgets, while the marginal improvements gradually diminish as $B_{\max}$ increases. Specifically, increasing $B_{\max}$ from $0$ to $1$ improves NarraCrime-Complex RVS by $+7.8$, EC by $+9.6$, and StrategyQA accuracy by $+12.3$, indicating that even a small refinement budget is sufficient to recover important evidence gaps and improve answer grounding. In contrast, when $B_{\max}$ increases from $3$ to $4$, the corresponding gains reduce to only $+1.3$, $+0.6$, and $+0.5$, respectively, showing a clear diminishing-return pattern at higher budgets. Overall, the main gains of EVAR are obtained in the low-to-medium budget regime, while larger budgets provide only limited additional benefit. This suggests that $B_{\max}=4$ offers a reasonable trade-off between performance and cost.

\subsection{Early-stopping Effectiveness}
\label{sec:exp_earlystop}

\begin{figure}[t]
    \centering
    \includegraphics[width=\linewidth]{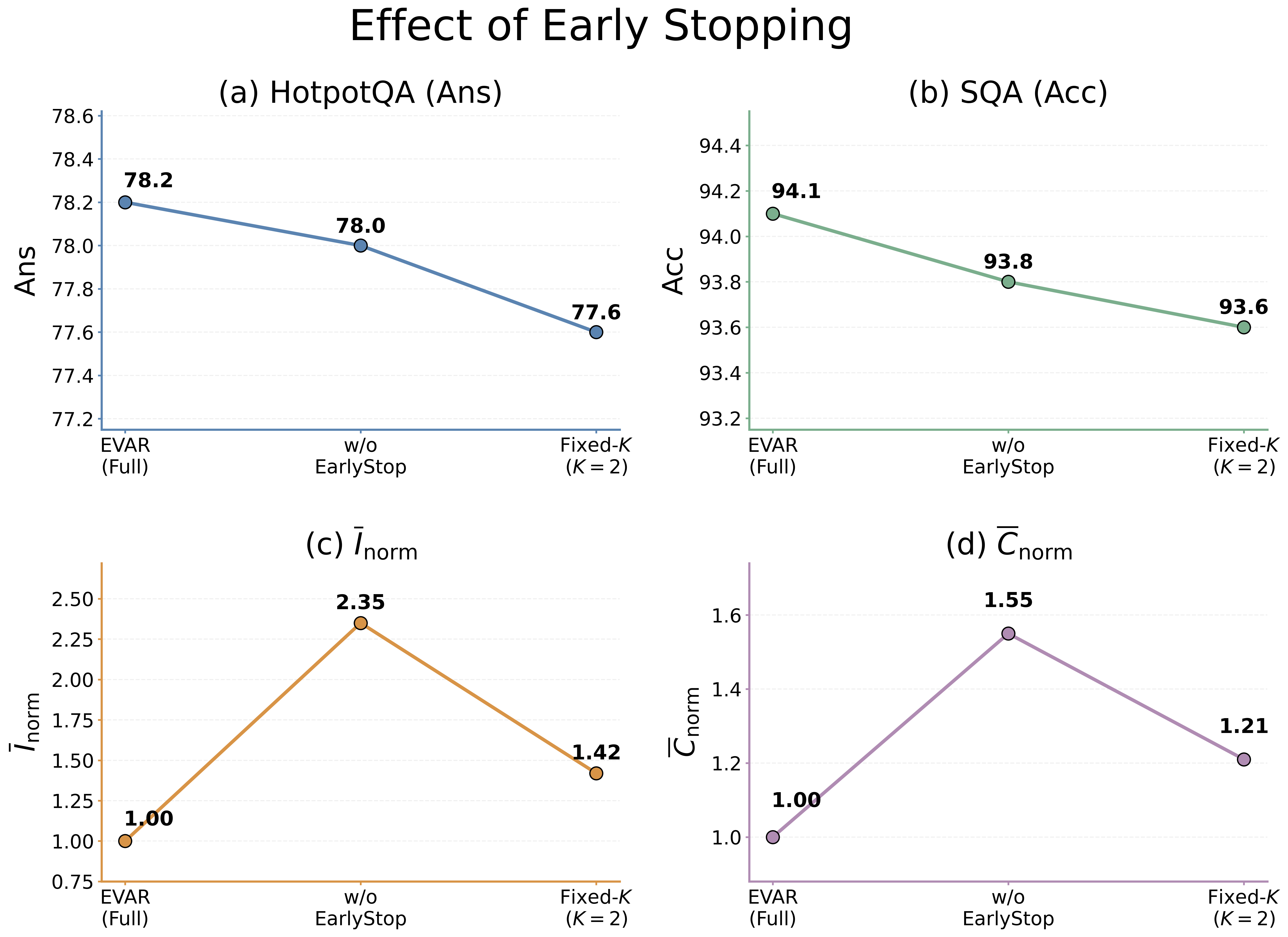}
    \caption{Early-stopping effectiveness of EVAR. Panels (a)--(d) show HotpotQA (Ans), SQA (Acc), $\overline{I}_{\mathrm{norm}}$, and $\overline{C}_{\mathrm{norm}}$, respectively, with EVAR (Full) normalized to 1.00.}
    \label{fig:earlystop_real}
\end{figure}

To quantify the effect of sufficiency-based early stopping, we compare EVAR (Full) with two variants: w/o EarlyStop and Fixed-$K$ ($K{=}2$). These variants keep the backbone model, prompts, routing, refinement operators, and admission mechanism unchanged, and differ only in the number of refinement iterations executed after routing to \textsc{Iter}. We report normalized efficiency as
\begin{equation}
\overline{I}_{\mathrm{norm}}=
\frac{\overline{I}}{\overline{I}^{\mathrm{Full}}},
\qquad
\overline{C}_{\mathrm{norm}}=
\frac{\overline{C}}{\overline{C}^{\mathrm{Full}}},
\label{eq:earlystop_norm}
\end{equation}
where $\overline{I}$ and $\overline{C}$ denote the average numbers of executed refinement iterations and LLM calls, respectively. For Fig.~\ref{fig:earlystop_real}, both averages are computed over the pooled HotpotQA and StrategyQA evaluation instances. As shown in Fig.~\ref{fig:earlystop_real}, EVAR (Full) achieves the best performance on both HotpotQA and StrategyQA, while its normalized iteration count and cost are both set to $1.00$. In contrast, w/o EarlyStop yields higher $\overline{I}_{\mathrm{norm}}$ and $\overline{C}_{\mathrm{norm}}$ but slightly worse performance, indicating that additional refinement brings extra cost without improving results. Fixed-$K$ ($K{=}2$) is more efficient than w/o EarlyStop, with both normalized metrics closer to 1, but it still underperforms EVAR (Full). Overall, these results show that sufficiency-based early stopping helps EVAR reduce unnecessary refinement while maintaining stronger overall performance.

\subsection{Additional Analysis}
\label{sec:additional_analysis}

We provide additional routing effectiveness and backbone robustness analyses in Appendix~\ref{app:extra_analysis}. The routing analysis shows that EVAR assigns iterative refinement more frequently to harder instances, while the backbone robustness analysis shows that EVAR remains effective across additional backbone models beyond the main DeepSeek-V3.2 setting.

\section{Conclusion}
\label{sec:conclusion}

We propose EVAR, a budget-aware test-time framework for non-interactive long-form narrative reasoning that grounds inference in a source-linked atomic evidence store and gates hypothesis admission through validation challenges. Experiments on NarraCrime and three public benchmarks show improved task performance and evidence grounding with controllable inference cost.

\section*{Limitations}

EVAR requires multiple LLM calls and is costlier than single-pass
prompting. Its performance depends on evidence extraction and
verification quality. The automatic evaluation further relies on
a GPT-5.5 judge for predicted-proposition extraction, atomic-claim
decomposition, and evidence-status labeling. Although the judge,
prompts, and decoding configuration are fixed across systems and
the judge is not shown method identities, evaluation errors and
model-specific preferences may affect the absolute IR, ASR, EC,
UCR, and CR scores, particularly when GPT-5.5 is itself evaluated
as a backbone. These metrics should therefore be interpreted as
fixed-judge automatic diagnostics rather than human-verified
assessments. Our experiments focus mainly on English benchmarks,
leaving multilingual, open-ended, and interactive settings for
future work.

\bibliography{custom}

\appendix

\section{Full EVAR Procedure}
\label{app:evar_algo}

\begin{table*}[!t]
\centering
\tiny
\setlength{\tabcolsep}{3.2pt}
\renewcommand{\arraystretch}{0.92}
\resizebox{0.88\textwidth}{!}{
\begin{tabular}{l l c c c c}
\toprule
Backbone & Method & RVS & EC & UCR$\downarrow$ & $T\downarrow$ \\
\midrule
\multirow{4}{*}{Qwen3-235B}
 & CoT         & 44.8 & 60.9 & 29.1 & 2.5  \\
 & Self-Ref.   & 53.2 & 65.4 & 26.0 & 6.4  \\
 & GoT         & 68.8 & 76.9 & 16.4 & 36.1 \\
 & EVAR        & \textbf{76.9} & \textbf{82.1} & \textbf{9.4}  & 16.4 \\
\midrule
\multirow{4}{*}{GPT-5.5}
 & CoT         & 50.2 & 65.1 & 25.8 & 2.2  \\
 & Self-Ref.   & 58.6 & 69.3 & 22.4 & 5.8  \\
 & GoT         & 72.6 & 79.5 & 13.8 & 33.9 \\
 & EVAR        & \textbf{82.4} & \textbf{86.0} & \textbf{7.1}  & 14.9 \\
\midrule
\multirow{4}{*}{Gemini-2.5}
 & CoT         & 49.1 & 64.3 & 26.5 & 2.3  \\
 & Self-Ref.   & 57.4 & 68.7 & 23.1 & 6.1  \\
 & GoT         & 70.9 & 78.8 & 14.5 & 34.6 \\
 & EVAR        & \textbf{80.7} & \textbf{84.9} & \textbf{7.6}  & 15.6 \\
\midrule
\multirow{4}{*}{Claude-4.5}
 & CoT         & 49.6 & 64.8 & 26.1 & 2.2  \\
 & Self-Ref.   & 57.9 & 69.0 & 22.7 & 5.9  \\
 & GoT         & 71.5 & 79.1 & 14.1 & 34.2 \\
 & EVAR        & \textbf{81.3} & \textbf{85.4} & \textbf{7.4}  & 15.1 \\
\bottomrule
\end{tabular}
}
\caption{Backbone robustness on NarraCrime-Complex with additional backbone models. For each backbone, EVAR is compared with CoT, Self-Refine, and GoT. RVS denotes Role-Aware Verdict Score, EC denotes Evidence Coverage, UCR denotes Unsupported Claim Rate, and $T$ denotes the average number of LLM calls per instance. Lower values are better for UCR and $T$; boldface marks the best predictive or faithfulness result within each backbone.}
\label{tab:backbone_robustness}
\end{table*}

\begin{table}[t]
\centering
\scriptsize
\setlength{\tabcolsep}{4pt}
\renewcommand{\arraystretch}{1.08}
\resizebox{\columnwidth}{!}{
\begin{tabular}{l l c}
\toprule
Dataset & Group & \%Iter \\
\midrule
\multirow{2}{*}{HotpotQA}
 & \#support = 2        & 33\% \\
 & \#support $>$ 2      & 68\% \\
\midrule
\multirow{3}{*}{StrategyQA}
 & Q-len Bottom 33\%    & 26\% \\
 & Q-len Middle 34\%    & 45\% \\
 & Q-len Top 33\%       & 71\% \\
\midrule
BBH
 & Harder subtasks avg. & 79\% \\
\bottomrule
\end{tabular}
}
\caption{Routing effectiveness on public reasoning benchmarks. \%Iter denotes the percentage of instances assigned a positive refinement budget and therefore routed to the iterative path.}
\label{tab:routing_public}
\end{table}

\begin{algorithm}[t]
\small
\caption{EVAR Inference Procedure}
\label{alg:evar}
\KwIn{Evidence store $\mathcal{B}$, goal $\mathcal{G}$, global budget cap $B_{\max}$}
\KwOut{Textual answer $\hat{y}$ and, for NarraCrime, a normalized verdict distribution $\mathbf{p}$}

$\mathcal{Z}_0 \leftarrow \mathcal{M}_{\pi_{\textsc{gap}}}(\mathcal{B}, \mathcal{G})$\;
Compute complexity score $\Gamma$ using Eq.~\ref{eq:complexity}\;
Compute refinement budget $K$ using Eq.~\ref{eq:budget}\;
Determine route $r$ using Eq.~\ref{eq:route}\;

\If{$r=\textsc{Fast}$}{
    $(\hat{y},\mathbf{p}) \leftarrow
    \mathcal{M}_{\pi_{\textsc{ans}}}
    \bigl((\mathcal{B},\emptyset),\mathcal{G}\bigr)$
    \tcp*[r]{NarraCrime}
    \Return $\hat{y}$ and, for NarraCrime, $\mathbf{p}$\;
}

Initialize $\mathcal{V}_{\textsc{hist},0} \leftarrow \emptyset$\;
Initialize $\mathcal{H}^{+}_{\textsc{hist},0} \leftarrow \emptyset$\;
Initialize $\mathcal{H}^{?}_{\textsc{hist},0} \leftarrow \emptyset$\;
Initialize $\mathcal{H}^{-}_{\textsc{hist},0} \leftarrow \emptyset$\;
$\mathcal{S}_0 \leftarrow
(\mathcal{B},\mathcal{V}_{\textsc{hist},0},
\mathcal{H}^{+}_{\textsc{hist},0})$\;

\For{$t=0,1,\dots,K-1$}{
    \If{$t>0$}{
        $\mathcal{Z}_t \leftarrow
        \mathcal{M}_{\pi_{\textsc{gap}}}
        (\mathcal{S}_t,\mathcal{G})$\;
    }

    \If{$\mathcal{Z}_t=\emptyset$}{
        \textbf{break}\;
    }

    $\mathcal{V}_t \leftarrow \emptyset$\;
    $\mathcal{H}^{+}_t \leftarrow \emptyset$\;
    $\mathcal{H}^{?}_t \leftarrow \emptyset$\;
    $\mathcal{H}^{-}_t \leftarrow \emptyset$\;

    \ForEach{$z \in \mathcal{Z}_t$}{
        $\mathcal{H}_{z,t} \leftarrow
        \mathcal{M}_{\pi_{\textsc{hyp}}}
        (z,\mathcal{S}_t)$\;

        \ForEach{$h \in \mathcal{H}_{z,t}$}{
            $\mathcal{V}_{h,t} \leftarrow
            \mathcal{M}_{\pi_{\textsc{chal}}}
            (h,\mathcal{B})$\;

            $\mathcal{V}_t \leftarrow
            \mathcal{V}_t \cup \mathcal{V}_{h,t}$\;

            $(\ell(h),\mathcal{E}_h) \leftarrow
            \mathcal{M}_{\pi_{\textsc{ver}}}
            (h,\mathcal{V}_{h,t},\mathcal{B})$\;

            \uIf{$\ell(h)=\textsf{Support}$}{
                $\mathcal{H}^{+}_t \leftarrow
                \mathcal{H}^{+}_t
                \cup \{(h,\mathcal{E}_h)\}$\;
            }
            \uElseIf{$\ell(h)=\textsf{Unknown}$}{
                $\mathcal{H}^{?}_t \leftarrow
                \mathcal{H}^{?}_t \cup \{h\}$\;
            }
            \Else{
                $\mathcal{H}^{-}_t \leftarrow
                \mathcal{H}^{-}_t \cup \{h\}$\;
            }
        }
    }

    $\mathcal{V}_{\textsc{hist},t+1} \leftarrow
    \mathcal{V}_{\textsc{hist},t} \cup \mathcal{V}_t$\;

    $\mathcal{H}^{+}_{\textsc{hist},t+1} \leftarrow
    \mathcal{H}^{+}_{\textsc{hist},t}
    \cup \mathcal{H}^{+}_t$\;

    $\mathcal{H}^{?}_{\textsc{hist},t+1} \leftarrow
    \mathcal{H}^{?}_{\textsc{hist},t}
    \cup \mathcal{H}^{?}_t$\;

    $\mathcal{H}^{-}_{\textsc{hist},t+1} \leftarrow
    \mathcal{H}^{-}_{\textsc{hist},t}
    \cup \mathcal{H}^{-}_t$\;

    $\mathcal{S}_{t+1} \leftarrow
    (\mathcal{B},
    \mathcal{V}_{\textsc{hist},t+1},
    \mathcal{H}^{+}_{\textsc{hist},t+1})$\;

    $\sigma_{t+1} \leftarrow
    \mathcal{M}_{\pi_{\textsc{suf}}}
    (\mathcal{S}_{t+1},\mathcal{G})$\;

    \If{$\sigma_{t+1}\ge\tau_{\textsc{suf}}$}{
        \textbf{break}\;
    }
}

Let $t^\star$ be the index of the terminal refinement state and
$\mathcal{S}_{t^\star}$ the corresponding state\;

$\mathcal{S}^{\mathrm{ans}}_{t^\star} \leftarrow
(\mathcal{B},
\mathcal{H}^{+}_{\textsc{hist},t^\star})$\;

$(\hat{y},\mathbf{p}) \leftarrow
\mathcal{M}_{\pi_{\textsc{ans}}}
(\mathcal{S}^{\mathrm{ans}}_{t^\star},\mathcal{G})$
\tcp*[r]{NarraCrime; otherwise return $\hat{y}$ only}

\Return $\hat{y}$ and, for NarraCrime, $\mathbf{p}$\;
\end{algorithm}

Algorithm~\ref{alg:evar} summarizes the complete inference procedure of EVAR after the evidence store has been constructed. Its central design principle is to separate candidate generation from state update: a hypothesis may be proposed during refinement, but it cannot influence answer synthesis until its evidential status has been explicitly verified. This separation prevents a plausible but unsupported intermediate statement from being repeatedly reused as if it were part of the original narrative.

Given the immutable evidence store $\mathcal{B}$, the reasoning goal $\mathcal{G}$, and the global budget cap $B_{\max}$, EVAR first probes the currently available evidence for unresolved gaps. Here, a gap refers to a missing or underspecified premise that blocks a sufficiently grounded answer, rather than to every detail absent from the narrative. The resulting gap set, together with the uncertainty and conflict annotations stored in $\mathcal{B}$, is used to compute the complexity score $\Gamma$ in Eq.~\ref{eq:complexity} and the instance-specific refinement budget $K$ in Eq.~\ref{eq:budget}. If $K=0$, the instance follows the \textsc{Fast} route and the answer is synthesized directly from the locked evidence store. If $K>0$, the instance enters the \textsc{Iter} route, where $K$ specifies the maximum number of refinement iterations. The budget is an iteration cap rather than the total number of LLM calls, because one iteration may contain multiple gaps, hypotheses, and verification operations, and early stopping may terminate the process before the cap is reached.

For iterative refinement, EVAR initializes the working state as $\mathcal{S}_0=(\mathcal{B},\mathcal{V}_{\textsc{hist},0},\mathcal{H}^{+}_{\textsc{hist},0})$. At iteration $t$, the gap operator re-examines the current state and identifies premises that still prevent a sufficient conclusion. Recomputing the gaps after each update is necessary because an admitted hypothesis may resolve an earlier gap or expose a different remaining dependency. Candidate hypotheses are then proposed directly for each unresolved gap. At this point, the candidates are only possible explanations: they have not yet been added to the answer-supporting state and are not treated as source evidence.

For every candidate $h$, EVAR constructs three hypothesis-conditioned validation challenges. The support challenge searches for evidence units that directly justify $h$; the counterevidence challenge checks whether any source-linked unit conflicts with $h$; and the prerequisite challenge checks whether $h$ depends on an indispensable premise that remains unsupported. These challenges make verification targeted to the particular factual commitments introduced by the candidate. They function as verification instructions rather than new evidence or new premises, and their wording cannot itself be used to support $h$. The challenge history is retained for audit and duplicate avoidance, but it is excluded from the answer-supporting state.

The verifier evaluates $h$ jointly with its validation challenges and the locked store $\mathcal{B}$, returning a label $\ell(h)$ and, when available, supporting evidence units $\mathcal{E}_h$. A hypothesis labeled \textsf{Support} is admitted together with $\mathcal{E}_h\subseteq\mathcal{B}$. A hypothesis labeled \textsf{Unknown} is placed in a quarantine buffer because the available evidence is insufficient to establish it, even if it appears plausible. A hypothesis labeled \textsf{Contradict} is discarded because it conflicts with the evidence. Only supported hypotheses are added to $\mathcal{H}^{+}_{\textsc{hist},t+1}$ and can affect later reasoning. Quarantined and contradictory hypotheses may be logged for diagnosis, but they remain outside $\mathcal{S}_{t+1}$. This admission rule blocks an unsupported proposal from becoming self-reinforcing merely because it was generated in an earlier iteration.

Refinement stops under any of three conditions: no blocking gap remains, the state-level sufficiency score reaches $\tau_{\textsc{suf}}$, or the assigned budget is exhausted. Candidate verification and state sufficiency serve different purposes: verification determines whether an individual hypothesis may enter the state, whereas sufficiency determines whether the admitted information is collectively adequate for answering $\mathcal{G}$. Let $\mathcal{S}_T$ denote the terminal refinement state. Before answer generation, EVAR forms $\mathcal{S}^{\mathrm{ans}}_T=(\mathcal{B},\mathcal{H}^{+}_{\textsc{hist},T})$. Consequently, the final answer can use the original evidence and admitted hypotheses with their evidence links, but cannot directly use quarantined hypotheses, contradictory hypotheses, or validation challenges.

Viewed end to end, the procedure has a one-way evidential flow: the locked store determines what can count as evidence, hypothesis generation supplies candidates, validation challenges specify how each candidate should be tested, and the verifier controls the only transition into the answer-supporting state. Later iterations may build on admitted hypotheses, but the provenance-preserving evidence store remains unchanged. EVAR can therefore refine difficult instances without allowing the refinement process to redefine its own evidential basis.

\section{Additional Analysis}
\label{app:extra_analysis}

The main experiments establish the overall effectiveness, evidence faithfulness, and cost of EVAR. We provide two complementary analyses to examine whether its observed behavior is consistent with the intended design. First, routing effectiveness tests whether the complexity-aware controller activates refinement more often for instances with stronger observable difficulty signals. Second, backbone robustness tests whether the gains of evidence-validated hypothesis admission persist across model families rather than depending on a favorable interaction with one backbone. These analyses distinguish the proposed control mechanism from the simpler explanations of uniformly adding computation or benefiting only from a particular model.

\subsection{Routing Effectiveness}
\label{app:routing_effectiveness}

A central objective of EVAR is to avoid allocating the same amount of computation to every instance. Applying the full loop uniformly would increase inference cost and expose already-solvable cases to unnecessary hypothesis generation. Conversely, assigning every example the same small budget could leave difficult cases under-refined. Ideally, instances that can already be resolved from the locked evidence store should follow the inexpensive \textsc{Fast} path, whereas instances containing unresolved gaps, uncertainty, or conflicts should be routed to \textsc{Iter}. We therefore analyze whether EVAR's routing decisions align with external indicators of reasoning difficulty on HotpotQA, StrategyQA, and BBH.

These public benchmarks do not provide difficulty labels directly comparable to the Easy, Medium, and Complex splits of NarraCrime. We consequently construct diagnostic groups using available dataset characteristics. For HotpotQA, examples are grouped by the number of annotated supporting facts, because integrating more facts generally requires a broader evidence chain. For StrategyQA, questions are divided into length-based quantiles. Question length is only a lightweight proxy, but it offers a reproducible indicator of the amount of information and compositional structure expressed in the input. For BBH, we report the average routing behavior on a predefined collection of harder subtasks. These group assignments are used only for post-hoc analysis and are not provided to EVAR as route labels.

For each difficulty group $g$, we compute the fraction of instances routed to iterative refinement:
\begin{equation}
\%Iter(g)
=
\frac{1}{|\mathcal{D}_g|}
\sum_{i\in\mathcal{D}_g}
\mathbb{I}[r_i=\textsc{Iter}],
\label{eq:iter_rate}
\end{equation}
where $\mathcal{D}_g$ is the set of instances in group $g$, $r_i$ is the selected route for instance $i$, and $\mathbb{I}[\cdot]$ is the indicator function. Since $r_i=\textsc{Iter}$ exactly when the assigned budget is positive, $\%Iter(g)$ measures how often EVAR judges that at least one refinement iteration is warranted. It should not be interpreted as the realized iteration count or inference cost: two iteratively routed examples may receive different budgets, generate different numbers of candidates, and stop at different times. The average call count $T$ reported elsewhere measures the actual computational cost.

As shown in Table~\ref{tab:routing_public}, the iterative-routing rate increases as the diagnostic difficulty signal becomes stronger. On HotpotQA, $\%Iter$ rises from $33\%$ for examples with exactly two supporting facts to $68\%$ for examples with more than two supporting facts, a difference of $35$ percentage points. On StrategyQA, the rate rises monotonically from $26\%$ in the shortest group to $45\%$ in the middle group and $71\%$ in the longest group. On the selected harder BBH subtasks, the average iterative-routing rate reaches $79\%$.

These results show that EVAR does not trivially send nearly every example to the same path. Simpler groups retain a substantial fast-path share, while groups that tend to require stronger evidence integration are much more likely to receive refinement. This behavior is consistent with the intended role of the gap and uncertainty signals in the complexity score. At the same time, the grouping variables are diagnostic proxies rather than complete ground-truth measures of difficulty: question length and supporting-fact count cannot capture every source of reasoning complexity. The analysis should therefore be interpreted as a routing sanity check, not as a claim that these proxies causally explain every individual decision.

The routing rate should not be read as a direct performance score. A higher rate is useful for a difficult group only when refinement improves evidence-grounded reasoning, while a lower rate is useful for a simple group only when direct synthesis remains sufficient. The main task-quality, faithfulness, and cost results provide this complementary evidence; Table~\ref{tab:routing_public} specifically shows that the controller changes its computational behavior across groups.

\subsection{Backbone Robustness}
\label{app:backbone_robustness}

The main experiments use DeepSeek-V3.2 as the default backbone. To test whether EVAR's gains generalize beyond this setting, we further evaluate the framework with Qwen3-235B, GPT-5.5, Gemini-2.5, and Claude-4.5 on NarraCrime-Complex. We select the Complex split because its longer narratives, denser evidence, and stronger cross-event dependencies create a demanding setting for both hypothesis generation and evidence verification.

For each backbone, we compare EVAR with CoT, Self-Refine, and GoT under the same task inputs and evaluation criteria. CoT represents conventional multi-step prompting without an explicit iterative state. Self-Refine adds iterative feedback and output revision but does not impose EVAR's evidence-based admission boundary. GoT is the strongest structured trajectory-based baseline in the main comparison, but it explores a substantially larger reasoning structure. We report Role-Aware Verdict Score (RVS), Evidence Coverage (EC), Unsupported Claim Rate (UCR), and the average number of LLM calls per instance ($T$). Considering these metrics jointly is important because a method may improve role-aware verdict quality by generating more content while also increasing unsupported claims or computational cost.

As shown in Table~\ref{tab:backbone_robustness}, EVAR achieves the highest RVS and EC and the lowest UCR for all four backbones. Its RVS ranges from $76.9$ to $82.4$, EC from $82.1$ to $86.0$, and UCR from $7.1$ to $9.4$. Relative to GoT, EVAR improves RVS by $8.1$--$9.8$ points and EC by $5.2$--$6.5$ points, while reducing UCR by $6.7$--$7.0$ points. EVAR also uses $14.9$--$16.4$ calls per instance, compared with $33.9$--$36.1$ for GoT, corresponding to approximately $55\%$ fewer calls. Thus, its gains cannot be explained by exploring more reasoning trajectories than the strongest structured baseline.

The absolute scores vary with backbone capability, but the direction and magnitude of EVAR's advantage remain stable. In particular, the UCR reduction persists together with the RVS and EC gains, showing that the improvement is not obtained by trading faithfulness for more aggressive generation. The lower call count relative to GoT further indicates that selective, verifier-gated refinement can be more efficient than expanding a substantially larger reasoning structure.

CoT and Self-Refine require fewer calls than EVAR, so EVAR is not the minimum-cost method in absolute terms. Their lower cost, however, is accompanied by substantially lower RVS and EC and higher UCR. EVAR occupies a different cost--quality operating point: it spends additional computation relative to simple prompting and revision, but uses that computation to test candidates before they alter the answer-supporting state. The consistency of this pattern across model families reduces the likelihood that the improvement is caused by a model-specific prompt artifact. It does not imply independence from backbone quality, because evidence extraction and verification still rely on the underlying model, but it shows that the benefit of verifier-gated admission is not confined to the default backbone.

\section{Details of Semantic Matching and Faithfulness Metrics}
\label{app:metric_details}

Our evaluation separates two complementary properties of a prediction. The first is coverage: whether the output recovers the intents, action schemas, and supporting evidence annotated in the reference. The second is faithfulness: whether the output introduces claims that cannot be supported by or that directly conflict with the annotated evidence. These properties should not be collapsed into one score. An output can cover most reference evidence while adding speculative details, or it can avoid unsupported claims by producing an incomplete answer. We therefore measure semantic recall with IR, ASR, and EC, and measure unsupported or conflicting additions separately with UCR and CR.

For each content type $X\in\{I,A,E\}$, let $\hat{\mathcal{P}}^{X}$ denote the propositions extracted from the model prediction and let $\mathcal{P}^{X}$ denote the corresponding gold reference propositions. The superscripts $I$, $A$, and $E$ refer to intent, action schema, and supporting evidence. Matching is performed within the same content type, so an intent proposition is compared with reference intents rather than with evidence propositions.

We encode every predicted and reference proposition using the \texttt{all-}\allowbreak\texttt{mpnet-}\allowbreak\texttt{base-v2} Sentence-Transformers encoder~\cite{Reimers2019SentenceBERT,Song2020MPNet}\footnote{\url{https://huggingface.co/sentence-transformers/all-mpnet-base-v2}} and compute cosine similarity for every predicted--reference pair. Exact string matching is inappropriate because two propositions can express the same action, intent, or evidential relation with different wording. A fixed threshold $\delta=0.8$ determines which pairs are semantically similar enough to be considered candidate matches. We then perform greedy one-to-one matching in descending order of similarity. A pair is accepted only if neither its predicted proposition nor its reference proposition has already been matched. This constraint prevents one broad prediction from receiving credit for several distinct reference propositions and prevents repeated paraphrases from repeatedly matching the same reference item.

The fixed threshold balances two errors. A substantially lower threshold could count topically related but semantically different propositions as matches, whereas an overly high threshold could reject valid paraphrases because of surface-form variation. Using the same $\delta$ and matching order for all systems ensures that differences arise from model outputs rather than method-specific evaluation tuning.

A reference proposition is counted as recovered when it belongs to an accepted pair. Proposition-level recall is computed as
\begin{equation}
\mathrm{Recall}(\hat{\mathcal{P}}^{X},\mathcal{P}^{X})
=
\frac{
\mathrm{match}(\hat{\mathcal{P}}^{X},\mathcal{P}^{X})
}{
|\mathcal{P}^{X}|
}.
\end{equation}
Matching is performed independently within each instance, and
IR, ASR, and EC are micro-averaged over the evaluation set by
summing accepted matches and reference propositions across instances.
IR applies this procedure to intent propositions,
ASR to action-schema propositions, and EC to supporting-evidence propositions. The same encoder, threshold, and one-to-one rule are applied to every
evaluated method. We use recall because these metrics
ask whether required reference content has been recovered.
They do not by themselves penalize extra generated
content; unsupported or contradictory additions are
measured separately by UCR and CR.

For the faithfulness metrics, each model output is first decomposed into atomic claims. Atomic decomposition is necessary because a single sentence may contain multiple factual commitments with different evidential status. Each claim is judged separately against the gold evidence annotations and assigned exactly one label: \textsf{Support}, \textsf{Unknown}, or \textsf{Contradict}. A claim is labeled \textsf{Support} when it is entailed or directly justified by the annotated evidence. It is labeled \textsf{Unknown} when the annotated evidence is insufficient to verify it, even if the claim appears plausible. It is labeled \textsf{Contradict} only when it conflicts with the annotated evidence.

UCR is the micro-averaged fraction of claims labeled either \textsf{Unknown} or \textsf{Contradict}, following Eq.~\ref{eq:ucr}. CR is the micro-averaged fraction labeled \textsf{Contradict}, following Eq.~\ref{eq:cr}. Thus, CR isolates the directly conflicting subset of unsupported claims, while \(\mathrm{UCR}-\mathrm{CR}\) represents claims that are unverifiable but not contradicted. Both denominators count all atomic claims generated across the evaluation set.

Lower UCR and CR indicate stronger evidence faithfulness. A large gap between UCR and CR indicates that unsupported content mainly consists of unverifiable speculation, whereas a CR close to UCR indicates that a large proportion of unsupported claims directly conflicts with the evidence. Because every contradicted claim is included in UCR, \(\mathrm{CR}\leq\mathrm{UCR}\) necessarily holds. Together with the task-quality metrics, these diagnostics distinguish final-answer quality, recovery of required evidence, and faithfulness of generated reasoning.

\end{document}